%% file: emnlp2021.tex
\pdfoutput=1
\documentclass[11pt,a4paper]{article}
\usepackage[]{emnlp2021}
\usepackage{times}
\usepackage{amsmath}
\usepackage{bm}
\usepackage{multirow}
\usepackage{latexsym}
\usepackage{subfigure}
\usepackage{graphicx}
\usepackage{amsfonts}
\usepackage{enumitem}
\usepackage[ruled,vlined]{algorithm2e}
\usepackage{algpseudocode}
\usepackage{makecell}
\usepackage{booktabs} % For formal tables

\usepackage{balance}

\usepackage{microtype}

\usepackage{xcolor}

\definecolor{orange}{RGB}{255, 154, 74}  
\definecolor{blue}{RGB}{76, 145, 205}
\definecolor{red}{RGB}{241,87,74}

\title{Learning to Selectively Learn for Weakly-supervised \\Paraphrase Generation}

\author{Kaize Ding$^1$\thanks{~~Work was done as an intern at Amazon Alexa AI.} \qquad ~Dingcheng Li$^{2}$ \qquad Alexander Hanbo Li$^{3}$ \\ {\bf Xing Fan$^2$} \qquad {\bf Chenlei Guo$^2$} \qquad {\bf Yang Liu$^2$} \qquad {\bf Huan Liu$^1$} \\

$^1$Arizona State University \qquad $^2$Amazon Alexa AI \qquad $^3$Amazon AWS AI\\

\texttt{\{kaize.ding,huan.liu\}@asu.edu} \\

\texttt{\{lidingch,hanboli,fanxing,guochenl,yangliud\}@amazon.com} \\}
\date{}

\begin{document}
\maketitle

\begin{abstract}

Paraphrase generation is a longstanding NLP task that has diverse applications for downstream NLP tasks. However, the effectiveness of existing efforts predominantly relies on large amounts of golden labeled data. Though unsupervised endeavors have been proposed to address this issue, they may fail to generate meaningful paraphrases due to the lack of supervision signals. In this work, we go beyond the existing paradigms and propose a novel approach to generate high-quality paraphrases with weak supervision data. Specifically, we tackle the weakly-supervised paraphrase generation problem by: (1) obtaining abundant weakly-labeled parallel sentences via retrieval-based pseudo paraphrase expansion; and (2) developing a meta-learning framework to progressively select valuable samples for fine-tuning a pre-trained language model, i.e., BART, on the sentential paraphrasing task. We demonstrate that our approach achieves significant improvements over existing unsupervised approaches, and is even comparable in performance with supervised state-of-the-arts.

% received much attention in the research community due to its broad applications, such as paraphrase generation, text summarization, and query rewriting. However, the success of existing efforts largely rely on cost-intensive labeled data. In practice, it is not always feasible to collect large amount of ground truth labels for training a powerful Seq2Seq model, rendering the neural Seq2Seq models less effective. On the contrary, we are able to collect abundant weakly labeled data in different ways (e.g., retrieval system or backtranslation) as complementary. However, the main challenge lies in how to determine each pair of sequences is useful or not for model training. In this work, we propose to adopt reinforcement learning to proactively select valuable training data for learning a better Seq2Seq language generation model. We demonstrate the effectiveness of the proposed framework for improving the model performance of se2seq models under the weak supervision setting. 

\end{abstract}

\section{Introduction}
Paraphrase generation is a fundamental NLP task that restates text input in a different surface form while preserving its semantic meaning. It serves as a cornerstone in a wide spectrum of NLP applications, such as question answering~\cite{dong2017learning}, machine translation~\cite{resnik2010improving}, and semantic parsing~\cite{berant2014semantic}. With the recent advances of neural sequence-to-sequence (Seq2Seq) architecture in the field of language generation, a growing amount of literature has also applied Seq2Seq models to the sentential paraphrasing task.

%  One potential solution is to leverage a small set of labeled data for enhancing the paraphrase generation performance.

% Thus, one can expect that the model would not be trained well and consequently, would not be able to generate diverse outputs.

Despite their promising results, collecting large amounts of parallel paraphrases is often time-consuming and requires intensive domain knowledge. Therefore, the performance of supervised methods could be largely limited in real-world scenarios. Due to this problem, unsupervised paraphrase generation has recently received increasing attention, but the development is still in its infancy. Generally, sampling-based or editing-based approaches~\cite{bowman2016generating,miao2019cgmh} fail to incorporate valuable supervised knowledge, resulting in less coherent and controllable generated paraphrases~\cite{liu2019unsupervised}. In this work, we propose going beyond the existing learning paradigms and investigate a novel research problem -- \textit{weakly-supervised paraphrase generation}, in order to push forward the performance boundary of sentential paraphrasing models with low-cost supervision signals. 

As an understudied problem, weakly-supervised paraphrase generation is challenging mainly because of the following reasons: 
\textbf{(i)} although weak supervision has been applied in different low-resource NLP tasks~\cite{dehghani2017neural,aprosio2019neural}, for paraphrase generation, it is unclear how to automatically acquire abundant weak supervision data that contains coherent, fluent and diverse paraphrases; \textbf{(ii)} weakly-labeled paraphrases tend to be noisy and are not equally informative for building the generation model~\cite{ren2018learning,li2019learning,yoon2020data}. Hence, selecting valuable parallel sentences from weakly-labeled data is vital for solving the studied problem; and \textbf{(iii)} the state-of-the-art paraphrasing methods are predominantly built upon traditional Seq2Seq models, while the necessity of learning from scratch largely magnifies the learning difficulty when dealing with scarce or noisy training data~\cite{guu2018generating}. Thus it is imperative to seek a more robust and knowledge-intensive backbone for learning with weakly-labeled paraphrases.

% bridge the gap between model performance and labeling-cost.

To address the aforementioned challenges, we present a novel approach for learning an effective paraphrasing model from weakly-supervised parallel data. By virtue of a simple yet effective \textit{pseudo paraphrase expansion} module, for each input sentence, we are able to obtain multiple similar sentences without unbearable labeling cost and treat them as paraphrases. To mitigate the inaccurate supervision signals within the weakly-labeled parallel data and build an effective paraphrasing model, we further select valuable parallel instances by proposing a novel framework named \textit{Learning-To-Selectively-Learn} (LTSL). Remarkably, LTSL leverages meta-learning to progressively exert the power of pre-trained language model, i.e., BERT~\cite{devlin2018bert} and BART~\cite{lewis2019bart}, with weakly-labeled paraphrasing data. From a meta-learning perspective, the BERT-based \textit{gold data selector} is meta-learned to select valuable samples from each batch of weakly paired sentences, in order to fine-tune and maximize the performance of the BART-grounded \textit{paraphrase generator}. Afterwards, the paraphrase generation performance change on a small validation set will be used to perform meta-optimization on the data selection meta-policy. This way the two pre-trained components \textit{gold data selector} and \textit{paraphrase generator} in LTSL are able to reinforce each other by continuously learning on a pool of meta-selection tasks. To summarize, the major contribution of this work is three-fold:

\begin{itemize}[leftmargin=*,noitemsep,topsep=1.5pt]
    \item We investigate an understudied research problem: weakly-supervised paraphrase generation, which sheds light on the research of sentential paraphrasing under a low-resource setting.
    
    \item We develop a framework LTSL, which is a new attempt of leveraging meta-learning to enhance pre-trained language model on paraphrase generation with costless weak supervision data.
    
    \item We conduct extensive experiments to illustrate the superiority of our approach over both supervised and unsupervised state-of-the-art methods on the task of paraphrase generation.

\end{itemize}

\begin{figure*}[t]
    \graphicspath{{figures/}}
    \centering
    \includegraphics[width=0.975\textwidth]{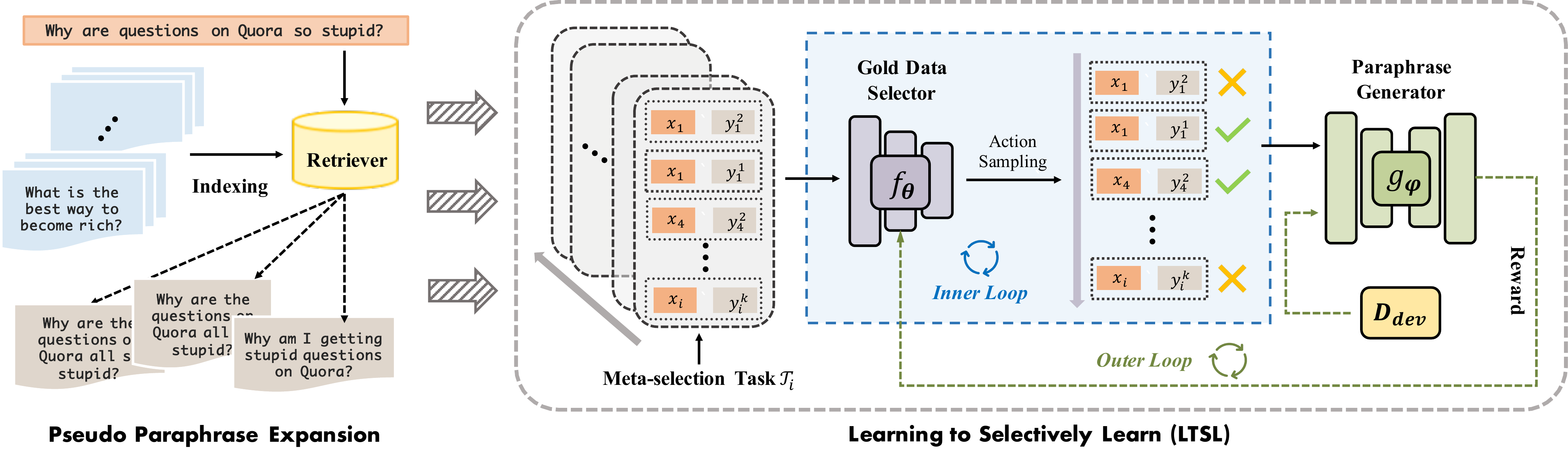}
    \caption{Overview of our approach for weakly-supervised paraphrase generation. For the LTSL framework, the blue dashed rectangle represents a meta-selection task, while green dashed line is the meta-optimization step.}%
    \label{fig:framework}%
\end{figure*}

% The generated examples can also be leveraged to train
% neural networks so that they become more robust
% to adversarial attacks (Iyyer et al., 2018). 

% The reason why not choose summarization:
% \begin{itemize}
%     \item If we use concatenated sentences as source and source as target, the objective of the model is to recover the source sentence. If we can learn a perfect summarizaiton model, the output of this model will always recover the source sentence, which is different from the actual goal of generating paraphrase.
    
%     \item For integrating summarization into the RL framework, the problem is that we need the validation set to provide signal. However, the validations set is paraphrase sentence pairs, which means the two tasks are inherently different, thus it cannot provide correct signals to the summarization. 
% \end{itemize}

\section{Related Work}

\noindent\textbf{Supervised Paraphrase Generation.} With the fast development of deep learning techniques, neural Seq2Seq models have achieved superior performance over traditional paraphrase generation methods that rely on exploiting linguistic knowledge~\cite{mckeown1980paraphrasing,mckeown1983paraphrasing} or utilizing statistical machine translation systems~\cite{dolan2004unsupervised,bannard2005paraphrasing}. Supervised paraphrasing methods are widely studied when the ground-truth parallel sentences are available during the training time. Among supervised efforts, Residual LSTM~\cite{prakash2016neural} is one of the earliest works based on neural networks. Later on, \citet{li2018paraphrase} propose to make use of deep reinforcement learning and \citet{iyyer2018adversarial,chen2019controllable} leverage syntactic structures to produce better paraphrases. More recently, retrieval-augmented generation methods have also been investigated~\cite{hashimoto2018retrieve,kazemnejad2020paraphrase,lewis2020retrieval} for paraphrase generation and achieved promising performance.

\smallskip
\noindent\textbf{Unsupervised Paraphrase Generation.} Due to the burdensome labeling cost of supervised counterparts, unsupervised paraphrasing methods have drawn increasing research attention in the community. Methods based on variational autoencoders (VAE) are first proposed to generate paraphrases by sampling sentences from the learned latent space~\cite{bowman2016generating,bao2019generating,fu2019paraphrase}, while the generated sentences are commonly less controllable. To tackle this issue, CGMH~\cite{miao2019cgmh} uses Metropolis-Hastings sampling to add constraints on the decoder at inference time. Furthermore, researchers try to improve the generation performance in terms of semantic similarity, expression diversity, and language fluency by using simulated annealing~\cite{liu2019unsupervised}, syntactic control~\cite{huang2021generating}, or dynamic blocking~\cite{niu2020unsupervised}. In addition, pre-trained translation models have been explored to generate paraphrases via back-translation~\cite{wieting2017learning,guoautomatically}. But still, those methods can hardly achieve comparable results with supervised approaches.

% Remarkably, our approach  different from both supervised and unsupervised methods and achieve state-of-the-art results with low-resource consumption. 

% As paraphrases capture the essence of language diversity, how to generate meaningful paraphrases has been studied over years in the research community.

%  Supervised \& Unsupervised. However, unsupervised methods. Guu et al. (2018) proposed the neural editor model for
% unconditional text generation, which produces a new sentence by editing a retrieved prototype using an edit vector. 

% Hashimoto et al. (2018) proposed a
% task-specific retriever using the variational framework to generate complex structured outputs. \citet{wu2019response} augmented Seq2Seq generation based models with retrieval frameworks to make the dialog responses more meaningful and nongeneric. \citet{lewis2020retrieval} explore a general-purpose fine-tuning recipe for retrieval-augmented generation models which combine pre-trained parametric and non-parametric memory for language generation. 

\smallskip
\noindent\textbf{Learning with Weak Supervision.}
The profound success of machine learning systems largely benefits from abundant labeled data, however, their performance has been shown to degrade noticeably in the presence of inaccurate supervision signals~\citep{hendrycks2018using}, especially in an adversary environment~\citep{reed2014training}. As one of the central problems in weak supervision, learning with noisy labels has received much research attention. Existing directions mainly focus on: estimating the noise transition matrix~\cite{goldberger2016training,patrini2017making}, designing robust loss functions or using regularizations~\cite{ghosh2017robust,li2017learning,zhang2020distilling}, correcting noisy labels~\cite{tanaka2018joint,zheng2021meta} and selecting or reweighting training examples~\cite{ren2018learning,chen2019understanding,yoon2020data}. In general, the state-of-the-art methods usually exploit a small clean labelled dataset that is allowed under the low resource setting~\cite{mirzasoleiman2020coresets}. For instance, Gold Loss Correction~\cite{hendrycks2018using} uses a clean validation set to recover the label corruption matrix to re-train the predictor model with corrected labels. Learning to Reweight~\citep{ren2018learning} proposes a single gradient descent step guided with validation set performance to reweight the training batch. Learning with noisy/weak supervision has drawn increasing attention in the NLP community~\cite{qin2018robust,feng2018reinforcement,ren2020denoising}, but it is seldomly investigated in the filed of paraphrase generation. In this work, we propose a new meta-learning framework that is capable of selecting valuable instances from abundant retrieved weakly-labeled sentence pairs.

\section{Proposed Approach}
Figure \ref{fig:framework} illustrates our method for solving weakly-supervised paraphrase generation. 
In essence, there are two sub-tasks: (1) \textit{how to obtain abundant weakly-labeled parallel data from the unlabeled corpus}; and (2) \textit{how to build a powerful paraphrase generation model from noisy weak supervision data}. 
Formally, given a set of source sentences $\mathcal{X} = \{x_i\}_{i=1}^{N}$ without ground-truth paraphrases, we first obtain a weakly-labeled parallel corpus $\mathcal{D}_{pseudo} = \{(x_i, y_i)\}_{i=1}^{M}$ for enabling weak supervision. In this work, we aim to denoise the weak supervision data by selecting a subset of valuable instances $\mathcal{D}_{train} = \{(x_i, y_i)\}_{i=1}^{M'}$ from $\mathcal{D}_{pseudo}$. 
A small set of trusted parallel sentences $D_{dev} = \{(x_i, y_i)\}_{i=1}^{L}$ ($L \ll M$) is allowed to be accessed, which is a common assumption in weakly-supervised learning~\cite{ren2018learning}. In the following subsections, we will introduce how to solve the main challenges with the proposed \textit{pseudo paraphrase expansion} module and the meta-learning framework LTSL.

% We first illustrate the way we . As suggested by Then we propose the meta-learning framework LTSL to .

% \begin{problem}
% \textbf{Weakly-supervised Paraphrase Generation}: , our goal can be decomposed to two sub-tasks: (1) ; and (2) learn a paraphrase generation model $G_{\bm\theta}$ that maximizes $p(y_i | x_i)$ using $\mathcal{D}_{pseudo}$.
% \end{problem}

% As shown in Figure 1, for each sentence, we first expand and compose 

%  our goal is to find the set of parameters of the model that maximizes
% To best utilize the information provided by the weak labels, we propose to construct a label correction network (LCN), serving as a meta model, which takes a pair of noisy data example and its weak label as input and produces a different version of the weak label.

% We consider curriculum learning on a pre-trained NMT
% model, where the goal is to improve an existing model
% pθ(y|x) by selecting a subset DS from the training dataset
% DT that led to pθ(y|x).

% Large collections of sentential paraphrase corpora could benefit such systems

\subsection{Pseudo Paraphrase Expansion}
% Pseudo
% To enable weakly-supervised paraphrase generation, we first propose a \textit{pseudo paraphrase expansion} module that obtains one or more pseudo (weakly-labeled) paraphrases that are similar or relative to an input sequence $x$.

To enable weakly-supervised paraphrase generation, we first propose a plug-and-play \textit{pseudo paraphrase expansion} module. Essentially, the function of this module is to obtain multiple weakly-labeled pseudo paraphrases that are similar or relative to each of the input sequence $x$.

% weakly labeled data 

\smallskip
\noindent\textbf{Expansion via Retrieval.} Inspired by the success of retrieval-enhanced methods~\cite{kazemnejad2020paraphrase,lewis2020retrieval} in text generation tasks, we propose to build a retrieval-based expansion module to obtain abundant pseudo parallel paraphrases $\mathcal{D}_{pseudo}$. 
Given a source sentence $x_i$, this module automatically retrieves a neighborhood set $\mathcal{N}(x_i)$ consisting of the $K$ most similar sentences $\{y_i^k\}^K_{k=1}$ from a large unlabeled sentence corpus. Specifically, we adopt the simple yet effective retriever BM25~\cite{robertson2009probabilistic} in this work. In addition, we use the Elastic Search~\cite{gormley2015elasticsearch} to create a fast search index for efficiently searching for the similar sentences to an input sequence. Here we use the in-domain sentence corpus since it is commonly available in practice and provides better results, but our approach is flexible to be extended to open-domain corpora such as Wikipideia.

% From our preliminary experiments, best results can be achieved by retrieving the top $k = 5$ sentences.

% BM25 is a probabilistic retrieval framework amd ranks a set of documents bd-overlapping, the similarity score between a query $Q$ and a document $d$ can be generally formulated as:
% \begin{equation}
%     Score(Q, d) = \sum_i^n W_i \cdot R(q_i, d),
% \end{equation}
% where $q_i$ is the $i$-th query term of $Q$ and $W_i$ represents the corresponding weight, which is commonly computed by $\text{IDF}(q_i)$. Here $R(q_i, d)$ measures the relevance score between $q_i$ and $d$. 

%  Note that using a pre-trained retriever can help us to alleviate
% the scarcity problem of the training data available
% for paraphrasing.

\smallskip
\noindent\textbf{Further Discussion.}
It is worth mentioning that the main reasons of using BM25 rather than a trainable retriever are: (1) this module is not only restricted to retrieval-based expansion, it is designed as a plug-and-play module that can provide more flexibility for weak supervision; and (2) the model training can be more stable since the number of trainable parameters is largely reduced. In addition to the aforementioned retrieval-based method, our approach is also compatible with other expansion alternatives. For instance, we can also adopt domain-adapted paraphraser to generate weakly-labeled paraphrases. Due to the simplicity and learning efficiency, here we focus on retrieval-based expansion to enable weakly-supervised paraphrase generation, and we leave the exploration of other expansion methods for future study.

% For instance, previous research~\cite{federmann2019multilingual} suggests that translation can be seen as paraphrasing the source sentence in a different language, thus it is natural to apply back-translation (e.g., ENG $\rightarrow$ FR $\rightarrow$ ENG) to acquire pseudo parallel data.

% , NMT is quite natural approach to paraphrasing in data augmentation
 
%  we observe that in translation, there is not a single correct translation target, but rather several variants of the sentence, carrying the same meaning, or paraphrases.

% An natural solution is to provide input in a foreign language. Nearly anything expressible in one human
% language can be written in another language. When
% users translate content, some variation in lexical
% realization occurs. To gather monolingual paraphrases, we can first translate a source sentence
% into a variety of target languages, then translate
% back into the source language, using either humans
% or machines. This provides naturalistic variation in
% language, centered around a common yet relatively
% unconstrained starting point. Although several research threads have explored this possibility (e.g.,
% (Wieting and Gimpel, 2018)), we have seen few if
% any comparative evaluations of the quality of this
% approach. they do produce
% greater diversity

\subsection{Learning to Selectively Learn (LTSL)}
The pseudo paraphrase expansion eliminates the dependency of a large amounts of ground-truth labels, nonetheless, one critical challenge is that the obtained weak supervision data is inevitably noisy: though a weakly-labeled paraphrase is somewhat related and convey overlapping information to the input sentence, while they are not parallel in the strict sense. As a result, directly using all the expanded pseudo paraphrase pairs for learning paraphrasing models is unlikely to be effective.

\smallskip
\noindent\textbf{Architecture Overview.}
To address the aforementioned challenge, we propose a meta-learning framework named Learning-To-Selectively-Learn (LTSL), which is trained to learn data selection meta-policy under weak supervision for building an effective paraphrasing model. LTSL consists of two components: (i) the meta-learner \textit{gold data selector} $f_{\bm\theta}(\cdot)$ parameterized by $\bm\theta$ that determines the selection likelihoods of the training samples to train the base model; and (ii) the base model \textit{paraphrase generator} $g_{\bm\phi}(\cdot)$ with parameters $\bm\phi$, which is a pre-trained autoregressive model that generates a paraphrase given the input sentence. At its core, the meta-learned gold data selector learns to select highly valuable samples from each batch, by measuring their ability to optimize the down-stream paraphrase generator. Meanwhile, the parameters of the paraphrase generator can be updated with the meta-selected samples progressively.

\smallskip
\noindent\textbf{Gold Data Selector (Meta-Learner).}
In order to represent each pair of parallel sentences, we adopt the widely recognized pre-trained model BERT~\cite{devlin2018bert} to build the gold data selector. Specifically, for the $i$-th weakly supervised paraphrase pair $(x_i , y_i)$, its latent representation can be computed by: 
\begin{equation}
    \mathbf{z}_i = \text{BERT}([\text{CLS}] \ x_i \  [\text{SEP}] \ y_i \  [\text{SEP}]),
\end{equation}
where [CLS] and [SEP] are special start and separator tokens. 
We use the last layer’s [CLS] token embedding as $\mathbf{z}_i$. 
Our gold data selector decides the value of the pair $(x_i, y_i)$ for fine-tuning the pre-trained paraphrase generator by:
\begin{equation}
    \mathbf{v}_i = \text{softmax}(\mathbf{W}_s\mathbf{z}_i + \mathbf{b}_s),
    \label{eqn:value}
\end{equation}
where both $\mathbf{W}_s$ and $\mathbf{b}_s$ are learnable parameters. Here $\mathbf{v}_i$ is the probability distribution of whether to include the weakly-labeled sentence pair $(x_i, y_i)$.
% we pack input sequence as ``[CLS] $x_i$ [SEP] $y_i$ [SEP]'',

% The rest of this section describes the State, Action, Reward, and the learning via Policy Gradient.

% The selected weak supervision pairs are then used to train the neural ranker to obtain the reward R, which goes back to train the action and state networks.

% By using the state representations $\mathbf{s}_i$ of the a-d pair.

\smallskip
\noindent\textbf{Paraphrase Generator.} The paraphrase generator could be built with any encoder-decoder backbones. In essence, the objective of the paraphrase generator is to maximize the conditional probability $p_{\bm\phi}(y|x)$ over the selected training samples in $\mathcal{D}_{train}$. As pre-trained language models are already equipped with extensive knowledge and have shown strong capability on a diverse set of generation tasks, we propose to use pre-trained language model BART~\cite{lewis2019bart} to build our paraphrase generator, which can largely reduce the difficulty of learning from scratch. Specifically, the fine-tuning objective of BART is:
\begin{equation}
    J(\bm\phi) = \sum_{(x,y) \in \mathcal{D}_{train}} - \log p_{\bm\phi}(y | x).
\end{equation}

% BART is pre-trained with a denoising objective, which endows the model with capabilities of both attending to the source sequence and paying attention to the context to generate coherent output.

% It has obtained state-of-the-art results .
% \begin{equation}
%     \log p_{\bm\theta}(y | x) = \sum_{i=1}^L log p_{\bm\theta} (y_i | y<i, x)
% \end{equation}
% Specifically, BART is a pre-trained Seq2Seq transformer that combines BERT and GPT. 
% The pre-training objective of BART is to reconstruct the original document from its corrupted version by an autoregressive decoder.

% We explore RAG models, which use the input sequence x to retrieve text documents z and use them
% as additional context when generating the target sequence y.

\smallskip
\noindent\textbf{Meta Reinforcement Learning.} 
Our proposed meta-learning framework LTSL aims to learn a discriminative data selection meta-policy for maximizing the performance of the paraphrase generator. For each batch of data samples in $\mathcal{D}_{pseudo}$, we consider it as a meta-selection task that contains a series of selective actions. As the data selection process is inherently non-differentiable, we adopt reinforcement learning (RL) to enable the meta-optimization.  We describe the RL environment of each meta-selection task as follows:

% The exploration of the data selection policy can be modeled by the meta-learned gold data selector.

\textsc{\textbf{State.}} The state $s_t$ is meant to be a summarization of the learning environment at time step $t$, which encodes the following information: (i) the representation of the $t$-th weakly-paired sentences; (ii) the
average of the representations of all selected sentences at time step $t$. The gold data selector will take the concatenated vector as input and output the probability distribution that indicates whether to select this instance or not.

% which is the representation $\mathbf{z}_t$ of the  in each sampled batch.

% Specifically, we take the hidden vector corresponding to [CLS] as the representation of current weakly paired instance $(x_i, y_i)$.

% It connects the selection policy and the environment.

\textsc{\textbf{Action.}} At each time step $t$, the action $a_t \in \{0, 1\}$ decides whether to select the current weakly-labeled instance $(x_t, y_t)$. Ideally, the gold data selector can take the action to select those useful instances for the paraphrasing task and filter out those noisy ones. Specifically, $a_t = 1$ represents the the current weakly-labeled instance will be selected, otherwise (i.e., $a_t = 0$) not. The action on each sentence pair is sampled according to the output of the selection policy $\pi_{\bm\theta}(\cdot | s_t)$.

% 

% we will omit the superscript k which denotes the bag index. Thus, the formulation hereafter is based on only one bag

% \begin{equation}
% \begin{aligned}
%     a_i &= \text{argmax}_{0,1} \pi(\mathbf{s}_i),\\
%     \pi(\mathbf{s}_i) &= \text{softmax}(\mathbf{W}\mathbf{s}_i + \mathbf{b}),
% \end{aligned}
% \end{equation}
% where the action probability $\pi(\mathbf{s}_i)$ is predicted with a simple linear layer on the state representation $\mathbf{s}_i$.

% \textsc{\textbf{Policy.}} The policy network (i.e., gold data selector) is trained to learn effective data selection meta-policy from weak supervision data. Essentially, the policy network receives a state representation $s_t$ then output the probability for each action $\pi_{\bm\theta}(\cdot | s_t)$ according to Eq.~\eqref{eqn:value}. By learning on a pool of meta-selection tasks, the meta-learned policy network is able to select valuable instances for enhancing the paraphrase generator.

\textsc{\textbf{Reward.}} 
The function of reward is to guide the meta-learner to select valuable training instances to improve the performance of the pre-trained generator. After each batch of selections, the accumulated reward of this meta-selection task is determined by the performance change of the paraphrase generator evaluated on the validation set $\mathcal{D}_{dev}$. Note that we use \textit{Perplexity} instead of word-overlapping metrics such as BLEU for evaluation since it is shown to be more efficient and stable~\cite{zhao2020reinforced} for generation tasks. For each meta-selection task, the policy network receives a delayed reward when it finishes all the selections, which is a commonly used design in RL literature~\cite{yao2019coacor}. 

% we assign a zero reward to intermediate selection steps

% can be obtained by quantify-ing the change of perplexity on the validation setafter the paraphrase generator is fine-tuned on theselected instances

% Specifically, the reward is determined by the performance change of the paraphrase generator evaluated on the validation set.
% \begin{equation}
%     R(\mathbf{s}_i, a_i) = PP(f(x_i,y_i)) - PP(fB(q,d))
% \end{equation}

% For the design of reward signal, we use the performance
% change of the paraphrase generator evaluated on the validation set.  We assign a reward of $0$ to unselected parallel pairs. 

\subsection{Meta-optimization}

To optimize the data selection meta-policy, we aim to maximize the sum of expected rewards of each meta-selection task, which can be formulated as:
\begin{equation}
    J(\bm\theta) = \mathbb{E}_{\pi_{\bm\theta}} [\sum_{t=1}^{T}r_t],
\end{equation}
where $r_t$ is the reward at time step $t$
and $\bm\theta$ is the parameter of the meta-learner gold data selector. We update the $\bm\theta$ via policy gradient:
\begin{equation}
\label{eq:update}
    \bm\theta^* \leftarrow \bm\theta + \alpha \nabla_{\bm\theta} \Tilde{J}(\bm\theta),
\end{equation}
where $\alpha$ denotes the learning rate. With the obtained rewards, the gradient can be computed by:
\begin{equation}
\label{eq:gradient}
    \nabla_{\bm\theta}  \Tilde{J}(\bm\theta) =  \sum_{t=1}^{B} r_t \nabla_{\bm\theta} \log \pi_{\bm\theta} (a_t|s_t),
\end{equation}
where $B$ is the number of instances in a batch. The details of the training process are shown in Algorithm \ref{alg:LTSL}. Specifically, we adapt the REINFORCE algorithm~\cite{williams1992simple} to optimize the policy gradients and implement a reward baseline to lower the variance during the training.

In essence, the obtained reward via a small validation set is used to conduct meta-optimization on the data selection meta-policy. By learning on a pool of meta-selection tasks, the meta-learned gold data selector can select valuable instances for enhancing the paraphrase generator. During the meta-learning process, the gold data selector and paraphrase generator can reinforce each other, which progressively improves the data selection policy and enhances the paraphrase generator.

\begin{algorithm}[t]
\caption{Learning algorithm of LTSL}
\label{alg:LTSL}
\LinesNumbered
\small
\KwIn{Weakly-labeled parallel set $\mathcal{D}_{pseudo}$ and the pre-trained language model $g_{\bm\phi}(y | x)$}
\KwOut{A paraphrase generation model $g_{\bm\phi}(y|x)$}

\While{$i < Epoch$}{
    // \texttt{Meta-selection Task $\mathcal{T}_i$}
    
    Sample $B$ samples $\mathcal{D}_B$ from $\mathcal{D}_{pseudo}$
    
    \For{$t = 1 \rightarrow B$} {
    Compute the state representation $s_t$
    
    Compute selection probabilities via Eq. (\ref{eqn:value})
    
    Sample the action $a_t$ for the current instance
    }
    
    // \texttt{Meta-optimization}
    
    Fine-tune the pre-trained generator $g_{\bm\phi}(y|x)$ with selected samples to get $g_{\bm\phi'}(y|x)$
    
    Calculate the reward $r_t$ on the validation set between $g_{\bm\phi}(y|x)$ and $g_{\bm\phi'}(y|x)$
    
    Update $\pi_{\bm\theta}$ according to Eq. (\ref{eq:update})
    
    \If{$i \bmod T == 0$}{
    Select data $D_{train}$ from $D_{pseudo}$ using $\pi_{\bm\theta}$
    
    Update the generator $g_{\bm\phi}(y|x)$ with $D_{train}$
    }
    
    }
    % \For{$j$ < $N$}{ 
    % $\mathcal{S}_i \leftarrow$ \textsc{Random}\textsc{Sample}($\mathcal{V}_i$, $N_S$)\;
    
    %  $\mathcal{Q}_i \leftarrow$ \textsc{Random}\textsc{Sample}($\mathcal{V}_i \backslash \mathcal{S}+i$, $N_Q$)\;

    \Return The fine-tuned paraphrase generator $g_{\bm\phi}(y|x)$\;
    
\end{algorithm}

\begin{table}[!b]
%\centering
\caption{Statistics of evaluation datasets}
\scalebox{0.9}{%
\begin{tabular}{lcccccc}
\toprule
 \textbf{Datasets} & \multicolumn{1}{c}{Train} & \multicolumn{1}{c}{Valid} & \multicolumn{1}{c}{Test} &
 \multicolumn{1}{c}{Corpus} &
 \multicolumn{1}{c}{Vocab} \\ \midrule
Quora-S & 100K & 3K & 30K & 400K & 8K  \\ 
Twitter &  110K & 1K & 5K & 670K & 8K\\ 
Quora-U &  117K & 3K& 20K & 400K & 8K\\ 
MSCOCO & 110K & 10K & 40K & 500K & 10K\\ 
\bottomrule
\end{tabular}
}
\label{table:dataset}
\end{table} 

\begin{table*}[t!]
\centering
\caption{Performance results of all the baseline methods on different paraphrasing datasets.}
\scalebox{0.805}{
\begin{tabular}{@{}l l ccccccccc@{}}
\toprule
\multicolumn{1}{c}{} & \multirow{2}{*}{\textbf{Method}}  &  \multicolumn{4}{c}{Quora-S}  & & \multicolumn{4}{c}{Twitter} \\ \cline{3-6} \cline{8-11} 

\multicolumn{1}{c}{} & \multicolumn{1}{c}{} & \multicolumn{1}{c}{\small BLEU-2} & \multicolumn{1}{c}{\small BLEU-4} & \multicolumn{1}{c}{\small ROUGE-1} &
\multicolumn{1}{c}{\small ROUGE-2}  & &

\multicolumn{1}{c}{\small BLEU-2} & \multicolumn{1}{c}{\small BLEU-4} & \multicolumn{1}{c}{\small ROUGE-1} &
\multicolumn{1}{c}{\small ROUGE-2}
\\ \midrule

\multirow{5}{*}{Supervised} & Res-LSTM   & $38.52$  & $24.56$  & $59.69$ & $32.71$ & & $32.13$  & $25.92$  & $41.77$ & $27.94$ \\

% & Pointer-generator &  $40.55$ & - & $61.96$   & $36.07$ & $30.21$ & & $40.37$ & - & $38.31$ & $21.22$ & $17.62$  \\

& Transformer & $42.91$  & $30.38$ & $61.25$ & $34.23$ & & $40.34$  & $32.14$ & $44.53$ & $29.55$   \\

& RbM & $43.54$ & -  & $38.11$ & $32.84$ & & $44.67$ & -  & $41.87$ & $24.23$ \\

& RaE & $40.35$  & $25.37$ & $62.71$ & $31.77$ & & $44.33$  & $34.16$ & $47.55$  & $31.53$ \\

& FSTE  & \textbf{51.03}  & $33.46$  & \textbf{66.17} & $39.55$ & & $46.35$  & $34.62$ & $49.53$ & $32.04$ \\
\midrule

\multirow{2}{*}{Weakly-supervised}& WS-BART  &  $44.19$ & $31.18$ & $58.69$  & $33.39$   & & $45.03$ & $34.00$ &  $51.34$ & $35.89$\\
%BART, Meteor, 58.69(Quora), 53.28(Twitter) 48.49 54.97

& LTSL (ours) & $49.18$  & \textbf{36.05} & $64.36$ & \textbf{39.71} & & \textbf{49.30} & \textbf{37.94} & \textbf{56.02} & \textbf{40.61} \\

\bottomrule

\end{tabular}}

\medskip

\scalebox{0.8}{
\begin{tabular}{@{}l l ccccccccc@{}}

\toprule
\multicolumn{1}{c}{} & \multirow{2}{*}{\textbf{Method}}  &  \multicolumn{4}{c}{Quora-U}  & & \multicolumn{4}{c}{MSCOCO} \\ \cline{3-6} \cline{8-11}

\multicolumn{1}{c}{} & \multicolumn{1}{c}{} & \multicolumn{1}{c}{\small \, iBLEU } & \multicolumn{1}{c}{\small \, BLEU \,} & \multicolumn{1}{c}{\small ROUGE-1} &
\multicolumn{1}{c}{\small ROUGE-2}  & &

\multicolumn{1}{c}{\small \, iBLEU} & \multicolumn{1}{c}{\small \, BLEU \,} & \multicolumn{1}{c}{\small \,ROUGE-1} &
\multicolumn{1}{c}{\small ROUGE-2}
\\ \midrule

\multirow{5}{*}{Unsupervised} 

% & VAE &  $8.16$ & $13.96$ & $44.55$  & $22.64$  & & $7.48$ & $11.09$ &  $31.78$ & $8.66$ \\

& CGMH  & $9.94$ & $15.73$ & $48.73$ & $26.12$  & & $7.84$ & $11.45$ &  $32.19$ & $8.67$    \\
& UPSA  & $12.02$ & $18.18$ & $56.51$  & $30.69$  & & $9.26$ & $14.16$ &  $37.18$ & $11.21$ \\
& PUP  &  $14.91$ & $19.68$ & $59.77$  & $30.47$  & & $10.72$ & $15.81$ &  $37.38$ & $13.87$  \\

& BackTrans  & $15.51$ & $26.91$ & $52.56$  & $27.85$  & & $7.53$ & $10.80$  & $36.12$  & $11.03$ \\
& set2seq+RTT  & $14.66$ & $22.53$ & $59.98$ & $34.09$  & & $11.39$ & $17.93$  & $40.28$  & $14.04$ \\
\midrule
\multirow{2}{*}{Weakly-supervised} & WS-BART  &  $17.04$ & $27.63$ & $56.43$  & $33.39$  & & $10.91$ & $15.90$ & $40.65$ & $15.62$\\
% FI-GCN-Con  & 77.5 & 77.2 && 54.7 & 53.8 && 73.4 & 72.1 && 
 & LTSL (ours) & \textbf{19.20}  & \textbf{29.25} & \textbf{61.71} & \textbf{39.21} & & \textbf{13.45} & \textbf{18.87} & \textbf{45.18} & \textbf{19.17}  \\

\bottomrule

\end{tabular}}

\label{table:all}
\end{table*}

% \begin{table*}[t!]
% \centering
% \caption{Comparison results with unsupervised models on two paraphrasing datasets.}

% \label{table:all}
% \end{table*}

\section{Experiments}
For the evaluation, we briefly introduce the experimental settings and conduct extensive experiments to corroborate the effectiveness of our approach. The details of our experimental settings and implementations can be found in the Appendix.

% , the corresponding source code and data are available at \url{https://github.com/ACL2021-LTSL/LTSL}.

\subsection{Experimental Settings}

\noindent\textbf{Evaluation Datasets \& Metrics.} In our experiments, we evaluate our proposed approach on multiple widely used paraphrasing benchmark datasets.
Since the problem of weakly-supervised paraphrase generation remains largely under-studied in the community, we compare our approach with both supervised and unsupervised paraphrase generation methods. 
%\paragraph{Evaluation Datasets \& Metrics.}
It is worth mentioning that, \textit{due to historical reasons, existing supervised and unsupervised methods use different data splits and evaluation metrics}. To make a fair and comprehensive evaluation, we follow the setting of each line of work and conduct the comparison respectively. Specifically, we use the following datasets to compare with supervised methods:

\begin{itemize}[leftmargin=*,noitemsep,topsep=1.5pt]
\item \textbf{Quora-S}: is the Quora question pair dataset which contains 260K non-parallel sentence pairs and 140K parallel paraphrases. Here we denote the version used by supervised methods as Quora-S. We follow the same setting in \citet{li2018paraphrase,kazemnejad2020paraphrase} and randomly sample 100K, 30K, 3K parallel sentences for training, test, and validation, respectively.

\item \textbf{Twitter}: is the twitter URL paraphrasing corpus built by \citet{lan2017continuously}. Following the setting in~\citet{li2018paraphrase,kazemnejad2020paraphrase}, we sample 110K instances from about 670K automatically labeled data as our training set and two non-overlapping subsets of 5K and 1K instances from the human-annotated data for the test and validation sets, respectively. 

% 670K (including parallel and non-parallel)

\end{itemize}
%  and additional 260K pairs of non-parallel sentences~\cite{liu2019unsupervised}
To compare our approach with unsupervised efforts, we adopt another two benchmark datasets:
\begin{itemize}[leftmargin=*,noitemsep,topsep=1.5pt]
    
    \item \textbf{Quora-U}: is the version of Quora dataset used by unsupervised paraphrasing methods. We follow the setting in ~\citet{miao2019cgmh,liu2019unsupervised} for a fair comparison and use 3K and 20K pairs for validation and test, respectively.
    
    \item \textbf{MSCOCO}: is an image captioning dataset containing 500K+ paraphrases pairs for over 120K image captions. We follow the standard splitting~\cite{lin2014microsoft} and evaluation protocols~\cite{liu2019unsupervised} in our experiments.

\end{itemize}

% To compare with unsupervised methods~\cite{liu2019unsupervised,siddique2020unsupervised}, we follow the standard train/test split, and randomly sample 10\% of the training data as the validation set.

The detailed dataset statistics are summarized in Table \ref{table:dataset}. Notably, although all the datasets have ground-truth paraphrases, our approach does not use them in the training set, which is as same as unsupervised methods~\cite{siddique2020unsupervised}. We only allow the model to access the parallel sentences in the validation set during the learning process. Specifically, when comparing with supervised baselines, we follow the previous works and adopt BLEU-n~\cite{papineni2002bleu} (up to n-grams), and ROUGE~\cite{lin2004rouge} scores as evaluation metrics; similarly, we use iBLEU~\cite{sun2012joint}, BLEU~\cite{post2018call} and ROUGE scores for comparing with unsupervised methods.

% More details about the evaluation datasets, metrics, and baselines can be found in Appendix.

%  For both two datasets, we set the vocabulary size to 8K as used in previous works.

% The methods we compare our approach to include the following:
\smallskip
\noindent\textbf{Compared Methods.} To show the superiority of our approach, we first include both widely used and state-of-the-art paraphrase generation methods as our baselines. In general, those methods can be divided into two categories: 
(1) \textit{supervised} methods that are trained with all the parallel sentences in the training corpus, including \textbf{Residual LSTM}~\cite{prakash2016neural}, \textbf{Transformer}~\cite{vaswani2017attention}, \textbf{RbM}~\cite{li2018paraphrase}, and two retrieval-based methods \textbf{RaE}~\cite{hashimoto2018retrieve} and \textbf{FSTE}~\cite{kazemnejad2020paraphrase}; 
(2) \textit{unsupervised} baselines without accessing ground-truth parallel data, including  \textbf{CGMH}~\cite{miao2019cgmh}, \textbf{UPSA}~\cite{liu2019unsupervised}, \textbf{PUP}~\cite{siddique2020unsupervised}, \textbf{BackTrans} (back-translation with pretrained NMT model) and  \textbf{set2seq+RTT}~\cite{guoautomatically}. 
%More details of those two categories of baselines can be found in Appendix A.2. 

% \textbf{VAE}~\cite{bowman2016generating},

In addition, we also include another \textit{weakly-supervised} method \textbf{WS-BART} where we use the same BART model~\cite{lewis2019bart} as in LTSL and directly fine-tune it with the clean validation set. Since this model is only fine-tuned with limited labeled data, here we consider it as a \textit{weakly-supervised} baseline.

\begin{figure}[t!]
    \graphicspath{{figures/}}
    \centering
    \scalebox{1.1}{
    \hspace{-0.3cm}
    \subfigure[\textbf{NDCG@K}] 
    {
    \includegraphics[width=0.475\columnwidth]{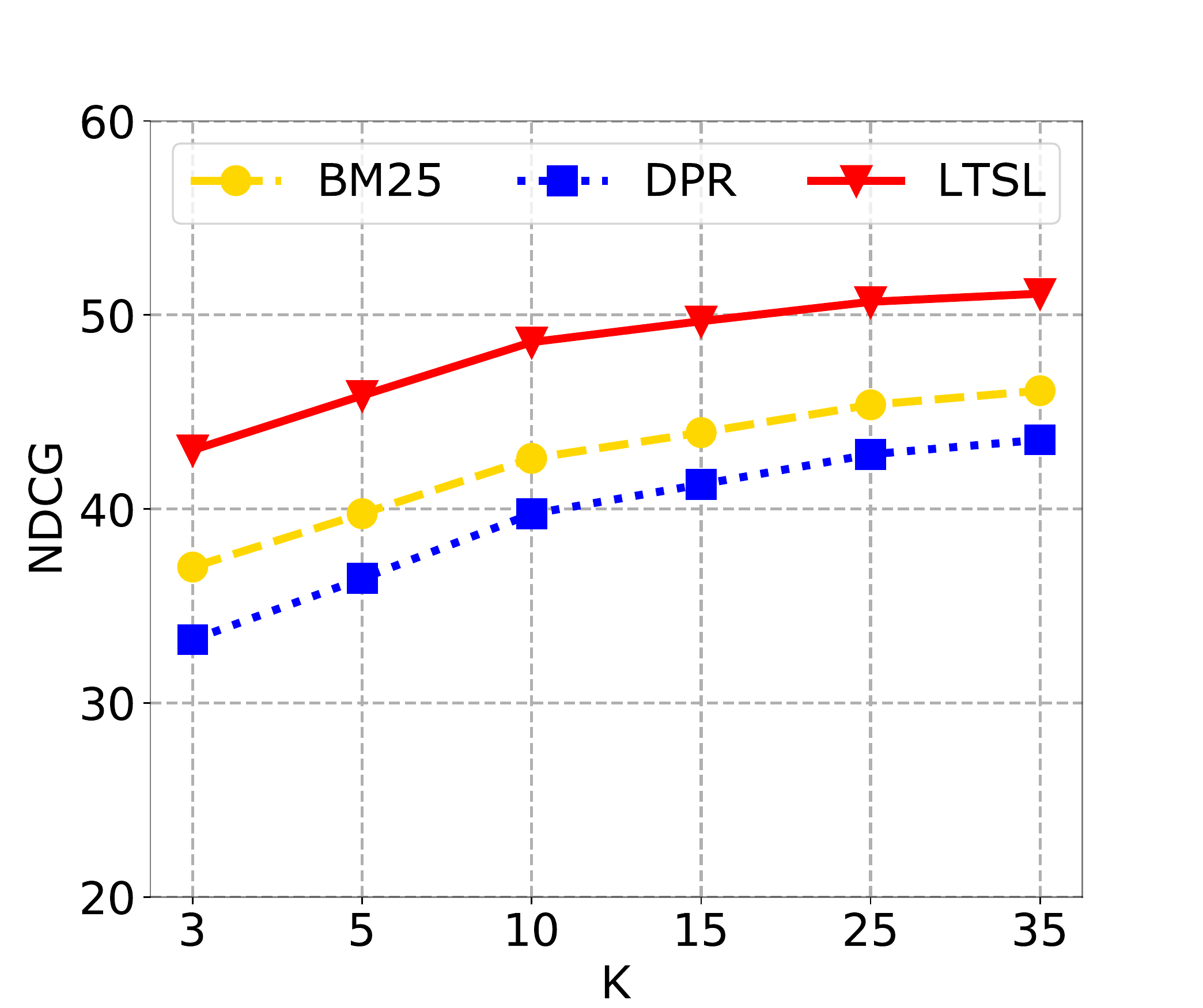}
    }
    \hspace{-0.5cm}
    \subfigure[\textbf{Recall@K}]
    {
    \includegraphics[width=0.475\columnwidth]{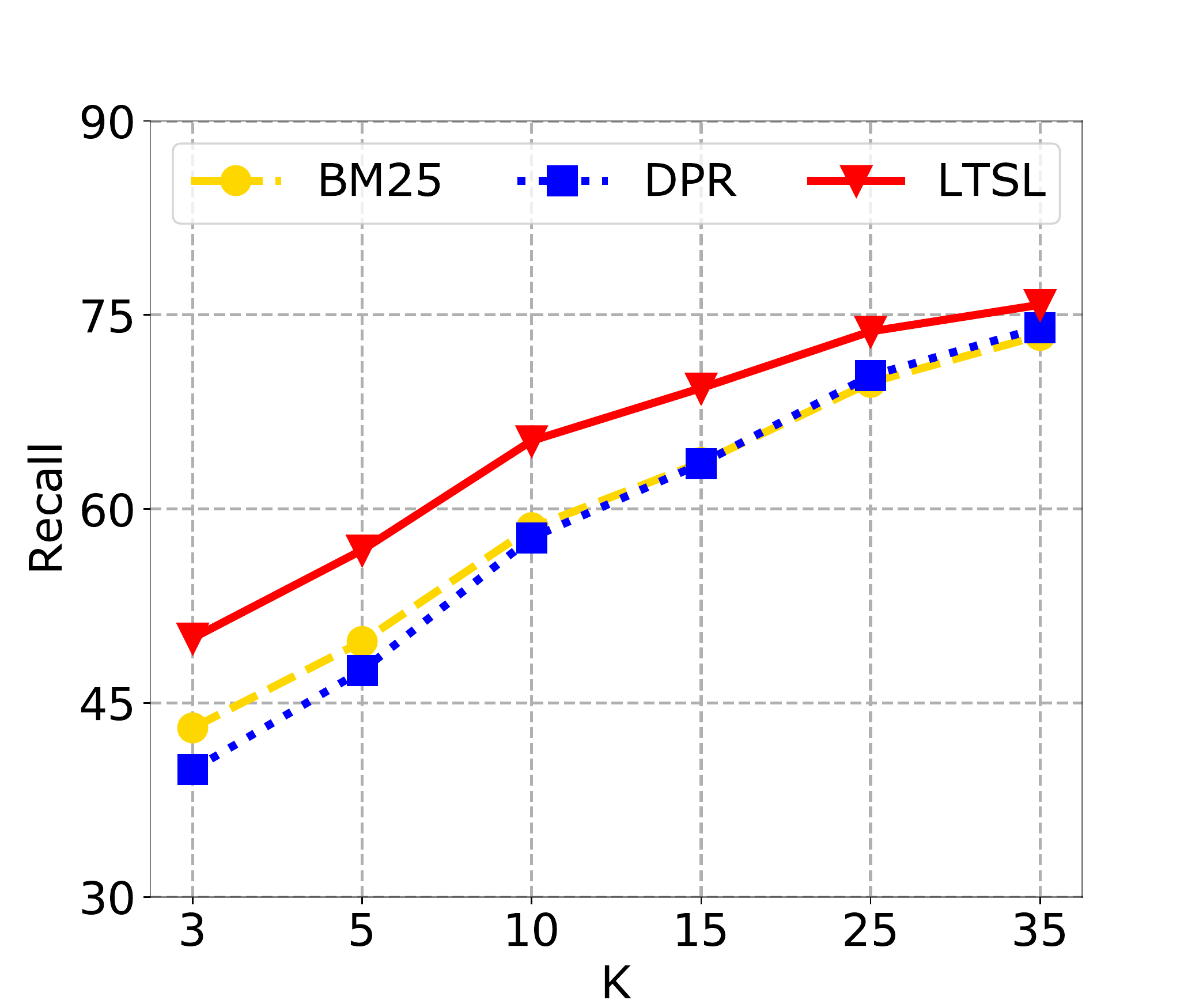}
    }
    }
    \caption{Selection effectiveness comparison results.}
    \label{fig:selection}
\end{figure} 

\subsection{Automatic Evaluation Results}
% \smallskip
\noindent\textbf{Paraphrase Generation.}
 Table \ref{table:all} summarizes the paraphrasing results of different methods. Overall, the results show that our approach LTSL achieves the state-of-the-art performance in most metrics. Specifically, we can make the following observations from the table:
\begin{itemize}[leftmargin=*,noitemsep,topsep=1.5pt]
    \item LTSL outperforms most of the \textit{supervised} baselines and achieves comparable performance to the state-of-the-art method (i.e., FSTE). In contrast with those supervised baselines that require large amounts of labeled parallel paraphrases, our approach LTSL delivers promising results with very low-cost supervision signals. This enables us to build an effective paraphrasing model under the real-world low-resource setting.
    
    % By virtue of the strength of the pre-trained language model, the generated paraphrase from pre-trained language models outperforms other methods by a significant margin.

    % Part of the performance gain of our model comes from the strength of the pretrained language model. 

    % However, FSTE requires large amounts of labeled parallel sentences, which is impractical in real-world scenarios and not generalizable to new domains.
    
    \item Compared to \textit{unsupervised} approaches, LTSL overall achieves large performance improvements, especially on  iBLEU and BLEU scores. The main reason is that the sampling or editing mechanisms in those methods lack  supervised knowledge from parallel sentences. As shown in 
    ~\cite{niu2020unsupervised}, those methods even fall behind a simple baseline \textit{copy-input} which directly copies the source sentence as the output. Our approach is able to well alleviate this weakness by leveraging the knowledge from weak supervision data and pre-trained language models. 
    
    % VAE achieve the worst performance on both datasets, indicating that paraphrasing by latent space sampling is worse than word editing. Part of the performance gain of our model comes from the strength of the pretrained language model, which achieves the top performance on
    % while obtaining the second best
    % ROUGE scores, indicating that these metrics fail
    % to punish against copying through the input. This
    % is consistent with observations from previous works ~\cite{mao2019polly,niu2020unsupervised}.
    
    \item By virtue of the strength of pre-trained language model, the \textit{weakly-supervised} baseline WS-BART performs relatively well compared to existing methods. However, it still falls behind our approach LTSL by a large margin, which demonstrates the effectiveness of our proposed learning framework.
\end{itemize}

\begin{table}[t!]
\centering
\caption{Ablation results on Quora-S dataset.}
\scalebox{0.8}{
\begin{tabular}{@{}l cccccc@{}}
\toprule
 \multirow{2}{*}{\textbf{Method}}  &  \multicolumn{4}{c}{Quora-S}  \\ \cline{2-5} 

& \multicolumn{1}{c}{\small BLEU-2} &  \multicolumn{1}{c}{\small BLEU-4} &  \multicolumn{1}{c}{\!\small ROUGE-1} & \multicolumn{1}{c}{\!\small ROUGE-2} 
\\ \midrule
pre-trained BART & $44.78$ & $31.44$ &  $55.35$ & $33.40$ \\

% w/o selection &  $33.02$   &  $34.88$ & $57.73$ \\

w/o selection & $44.15$  & $31.98$ &  $55.64$ & $33.65$ \\

w/o BART & $45.79$  & $32.18$ &  $57.69$ & $35.39$ \\

% Ret ($K=1$) + BART &  $33.55$   &  $35.50$ & $58.91$ \\
LTSL  &  \textbf{49.18}  & \textbf{36.05} &  \textbf{64.36} & \textbf{39.71}\\

\bottomrule

\end{tabular}}
\label{table:ablation}
\end{table}

\smallskip
\noindent\textbf{Gold Data Selection.}
To further evaluate the effectiveness of the meta-learned data selection policy, we use the well-trained gold data selector to compute data values on unseen weakly-supervised candidates during training. Specifically, for each sentence in the test set, we retrieve 50 most similar sentences and consider them as weakly-labeled paraphrases, and then the gold data selector is used to predict the quality of these paraphrase candidates. 
Note that here we also include the adopted retriever, BM25, and a BERT-based Dense Passage Retriever (DPR) fine-tuned with the labeled parallel sentences from validation set for comparison. 
We use the computed probabilities to rank all the candidates and report NDCG@K and Recall@K in Figure \ref{fig:selection} (a) and (b), respectively. 
We can see from the figures that the meta-learned gold data selector from LTSL is able to better select ground-truth parallel data than BM-25 and DPR. This observation indicates that the meta-learned gold data selector can effectively generalize to unseen data and select valuable weakly-labeled instances.

\begin{figure}[b!]
    \graphicspath{{figures/}}
    \centering
    \includegraphics[width=1.0\columnwidth]{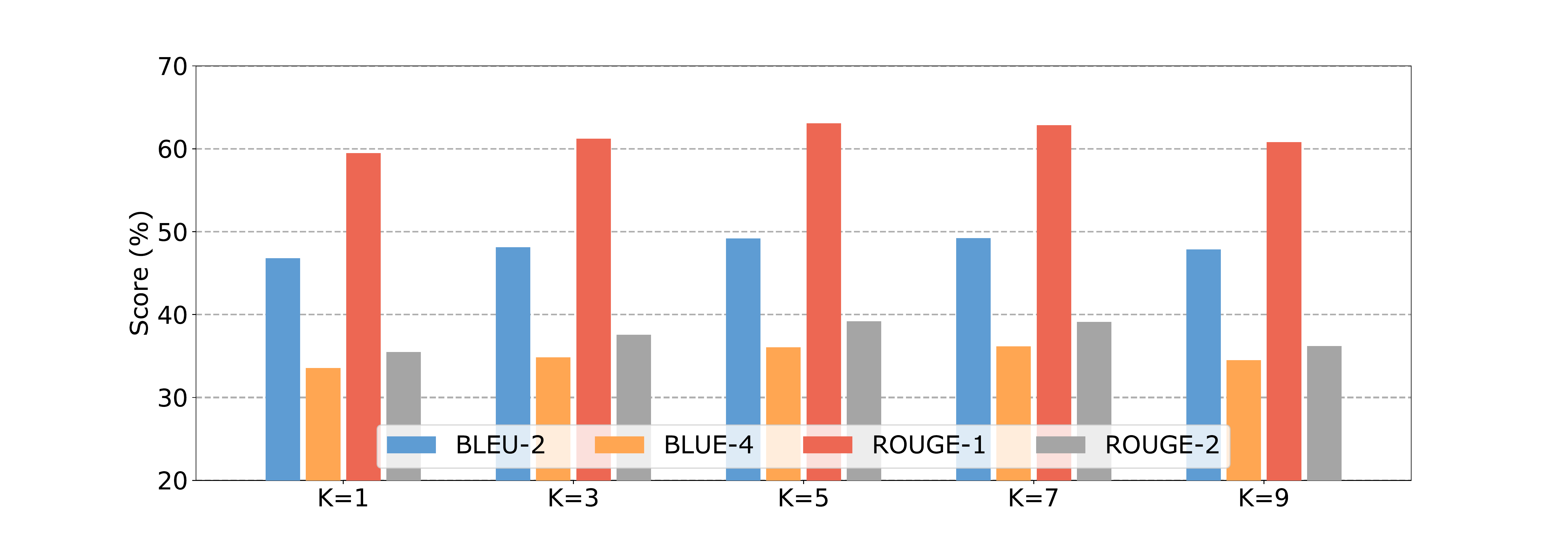}
    \caption{Parameter analysis ($K$) on Quora-S dataset.}%
    \label{fig:parameter}%
\end{figure}

\begin{table*}[t!]
\centering
\scriptsize
\caption{Examples of the generated paraphrases on Quora-U dataset. We highlight the key phrases in the paraphrases generated by each method, and we use underline to show the matched parts between LTSL and reference.}
\scalebox{1.0}{
\begin{tabular}{@{}c ccccc@{}}
\toprule

Input & BackTrans & WS-BART &  LTSL & Reference
\\ \midrule

\makecell[l]{How do I attract contributors \\for my project on Github?}  &  \makecell[l]{How do I \textcolor{orange}{attract contributors} \\for my Github project?} &  \makecell[l]{How do I \textcolor{red}{get my project} \\ \textcolor{red}{up and running} on Github?}   &  \makecell[l]{How do I \textcolor{blue}{\underline{get more} }\\\textcolor{blue}{\underline{contributors}} on my GitHub?} & \makecell[l]{How do I \underline{can get more} \\\underline{contributors} on GitHub?} \\
\midrule

% \makecell[l]{How do I to score good \\marks in 12th boards?} &  \makecell[l]{How do I to score marks in \\ in 14th boards?}  & \makecell[l]{ How do I score marks in \\\textcolor{black}{geography in ICSE} boards?}  &  \makecell[l]{How do I score good marks \\in class 12th board exams?} & \makecell[l]{How do I score good marks\\in class 12 board exams?} \\
% \midrule
\makecell[l]{Is Azusa Pacific University \\accepting LGBT people?} &
\makecell[l]{\textcolor{orange}{Does} Azusa Pacific \\ University \textcolor{orange}{accept} \\ \textcolor{orange}{LGBT people}?}  &
\makecell[l]{\textcolor{red}{How accepting is} Azusa \\Pacific University of \\\textcolor{red}{LGBT people}?}  &\makecell[l]{\textcolor{blue}{\underline{Should I worry about}} \\\textcolor{blue}{\underline{attending}} Azusa Pacific \\University \textcolor{blue}{\underline{if I am} LGBT}?}  & \makecell[l]{\underline{Should I worry about} \\\underline{attending} Azusa Pacific \\ University \underline{if I am gay}?}  \\
\midrule

\makecell[l]{What are the strangest \\ facts about some famous \\Bollywood movies?} &  \makecell[l]{What are \textcolor{orange}{the strangest facts} \\ about some famous \\Bollywood movies?} & \makecell[l]{What are the \textcolor{red}{strangest} \\\textcolor{red}{or weirdest facts} \\ about Bollywood movies?} &  \makecell[l]{What are \textcolor{blue}{\underline{some of the}}\\ \textcolor{blue}{\underline{strangest facts}} about \\ famous Bollywood movies?} & \makecell[l]{What are \underline{some of the} \\\underline{strangest facts} about \\ famous Bollywood movies?}\\

\midrule
\makecell[l]{What could be the reason \\behind Arnab Goswami \\quitting Times Now?} &
\makecell[l]{What \textcolor{orange}{can be the reason} \textcolor{orange}{of} \\ Arnab Goswami \textcolor{orange}{quitting} \\ Times Now?}  &
\makecell[l]{What \textcolor{red}{is the reason of} \\ Arnab Goswami \textcolor{red}{leaving} \\ Times Now?}  &\makecell[l]{\textcolor{blue}{\underline{Why} did} Arnab Goswami\\ \textcolor{blue}{\underline{resign}} from Times Now?}  & \makecell[l]{\underline{Why} Arnab Goswami \\\underline{resigned} Times Now?}  \\

\bottomrule

\end{tabular}}
\label{table:case}
\end{table*}

\smallskip
\noindent\textbf{Ablation \& Parameter Study.}
Next, we conduct ablation experiments to investigate the contributions of each component in the proposed approach. From the results reported in Table \ref{table:ablation}, we can learn that by removing the gold data selector, the variant \textit{w/o selection} only achieves similar results with \textit{pre-trained BART} (a variant without pseudo paraphrase expansion and gold data selection). It again verifies the importance of learning effective data selection meta-policy. Meanwhile, by using a vanilla Transformer to build the paraphrase generator (w.r.t., \textit{w /o BART}), the performance falls behind LTSL by a considerable gap, which shows the necessity of leveraging pre-trained language models.

% the performance of BART could be improved by doing retrieval-based pseudo paraphrase expansion. In the meantime, our approach LTSL achieves considerable improvements, which again demonstrates that the  gold data selector is able to select valuable instances for building a better paraphrase generator. \textit{Pre-trained BART} can be considered as a variant of our approach without pseudo paraphrase expansion and gold data selection; and the other three variants (e.g., \textit{Ret (K=1) + BART}) are BART models that directly fine-tuned with different size of retrieval-expanded parallel data.

We also examine the effect of parameter $K$ on the final performance and show the results on the Quora-S dataset in Figure \ref{fig:parameter} (similar results can be observed for other datasets). As we can see from the results, with the growth of $K$, the performance reaches the peak when $K$ is set to $5$ and then gradually decreases if $K$ increases. It shows that it is necessary to incorporate abundant weakly-labeled data. However, when more candidates are added, the introduced noise from weakly-labeled data could impair the final performance.

\begin{table}[h!]
\centering
\caption{Human evaluation results on Quora-U dataset.}
\scalebox{0.8}{
\begin{tabular}{@{}l cc c cc c cc@{}}
\toprule
 \multirow{2}{*}{\textbf{Method}}  &  \multicolumn{2}{c}{Coherence} &  & \multicolumn{2}{c}{Fluency} &  & \multicolumn{2}{c}{Diversity}\\ \cline{2-3} \cline{5-6} \cline{8-9} 

&  \multicolumn{1}{c}{Score\!} &  \multicolumn{1}{c}{\!$\kappa$\!} & & \multicolumn{1}{c}{Score\!} &  \multicolumn{1}{c}{\!$\kappa$\!} & & 
\multicolumn{1}{c}{Score\!} &  \multicolumn{1}{c}{\!$\kappa$}
\\ \midrule
BackTrans &  $3.86$ &  $0.44$  & & $3.95$ & $0.48$ && $3.09$ & $0.38$\\

WS-BART &  $4.04$   &  $0.49$ && $4.60$ & $0.51$ && $3.25$ & $0.43$ \\

LTSL  &  \textbf{4.39}  & \textbf{0.57} && \textbf{4.84} & \textbf{0.63} && \textbf{3.54} & \textbf{0.48} \\

\bottomrule

\end{tabular}}
\label{table:human}
\end{table}

\subsection{Human Evaluation Results}
To further illustrate the superior quality of the paraphrases generated by LTSL, we conduct subjective human evaluations. We randomly select 100 sentences from the Quora-U dataset and ask three human annotators to evaluate the top three performing methods under the unsupervised paraphrase generation setting. Table \ref{table:human} presents the average scores along with the inter-annotator agreement (measured by Cohen's kappa $\kappa$) in terms of semantic coherence, language fluency, and expression diversity. We rate the paraphrases on a scale of 1-5 (1 being the worst and 5 the best) for the three evaluation criteria.  As shown in the table, our approach outperforms all the competing approaches in terms of all the three perspectives. Moreover, the inter-annotator agreement shows moderate or good agreement between raters when assessing the outputs of our model.

% \footnote{We do not compare with PUP or DBlock in our human evaluation and case study since their codes are unavailable.}.

\subsection{Case Study}
Last, we showcase the generated paraphrases from different methods on the Quora-U dataset. As illustrated in Table~\ref{table:case}, we can clearly see qualitatively that LTSL can produce more reasonable paraphrases than the other methods in terms of both closeness in meaning and difference in expressions. For example, ``How do I attract contributors for my project on Github?'' is paraphrased as ``How do I get more contributors on my GitHub project?''. It is worth mentioning that existing unsupervised methods such as 
BackTrans cannot generate high-quality paraphrases in terms of diversity, mainly because of the shortage of supervision signals. On the contrary, our LTSL approach is able to generate highly fluent and diverse paraphrases by leveraging valuable weak supervision data and the knowledge of large-scale pre-trained language model.

% In the meantime, we can observe that the generated sentences from LTSL are when comparing to unsupervised methods . 
% One reasonable explanation is that conventional unsupervised methods are based on , which can hardly 
% Admittedly, such samples are relatively rare, and our current UPSA mainly synthesizes paraphrases by editing words in the sentence, whereas the syntax is mostly preserved. 

% This is partially due to the difficulty of exploring the entire (discrete) sentence
% space even by simulated annealing, and partially
% due to the insensitivity of the similarity objective
% given two very different sentences.

% Table 4 shows some examples of the paraphrases
% generated by our model. A common pattern among
% the output paraphrases is that the model has combined different parts of the input sentence and the
% retrieved target sentence to create a grammatically
% correct paraphrase.

\section{Conclusion}
In this work we investigate the problem of paraphrase generation under the low-resource setting and propose a weakly-supervised approach. From automatic and human evaluations, we demonstrate that our approach achieves the state-of-the-art results on benchmark datasets. An interesting direction is to improve the generation performance by leveraging weakly-labeled data from different sources. We leave this as future work.
%  By composing abundant weakly-labeled parallel data from a retriever (i.e., BM25), we further propose a meta-learning framework LTSL to progressively learn data selection meta-policy for fine-tuning the pre-trained language model.

% \paragraph{Sensitivity Analysis.}
% We examine the performance change in terms of the size of validation set.
\newpage

\bibliographystyle{acl_natbib}
\bibliography{reference}

\newpage
\appendix
\input{Appendix.tex}

\end{document}

%% file: Appendix.tex
\section{Appendix}

\subsection{Evaluation  Details}
For all the evaluation datasets, we follow previous research to pre-process the datasets. For the comparison with supervised methods, we adopt two benchmark datasets (i.e., Quora-S\footnote{\url{https://www.kaggle.com/c/quora-question-pairs}} and Twitter\footnote{\url{https://languagenet.github.io/}}) and truncate sentences in both of the datasets to 20 tokens as in \citet{li2018paraphrase,li2019decomposable}. For the comparison with unsupervised methods, we use Quora-U and MSCOCO\footnote{\url{https://cocodataset.org/\#home}}. Due to the space limit and data quality issues~\cite{niu2020unsupervised}, other datasets such as WikiANswers, Twitter (unsupervised version) are not included in our evaluations. We follow their settings and all the sentences are lower cased, and truncate all
sentences to up to 20 words. 

Throughout the paper, we use those evaluation metrics that have been widely used in the previous work to measure the quality of the paraphrases. In general, BLEU measures how much the words (and/or n-grams) in the machine generated summaries appeared in the human reference summaries. Rouge measures how much the words (and/or n-grams) in the human reference summaries appeared in the machine generated summaries. Specifically, we use the library\footnote{\url{https://huggingface.co/metrics/sacrebleu}} from HuggingFace to compute BLEU scores and \textit{py-rouge}\footnote{\url{https://pypi.org/project/py-rouge/}} to compute ROUGE scores. As BLEU and ROUGE could not measure the diversity between the generated and the original sentences, we follow unsupervised paraphrasing methods and adopt iBLEU~\cite{sun2012joint} to measure the diversity of expression in the generated paraphrases by penalizing copying words from input sentences. Specifically, we follow the unsupervised paraphrase generation baselines and set
the balancing parameter $\alpha = 0.9$. 

% For the adopted , we follow Li et al. (2019) to report iBLEU (Sun and Zhou, 2012) and ROUGE (Lin,
% 2004) on QQP.

% \begin{table}[!h]
% %\centering
% \caption{Statistics of evaluation datasets}
% \scalebox{0.9}{%
% \begin{tabular}{lcccccc}
% \toprule
%  \textbf{Datasets} & \multicolumn{1}{c}{Train} & \multicolumn{1}{c}{Valid} & \multicolumn{1}{c}{Test} &
%  \multicolumn{1}{c}{Corpus} &
%  \multicolumn{1}{c}{Vocab} \\ \midrule
% Quora-S & 100K & 3K & 30K & 400K & 8K  \\ 
% Twitter &  110K & 1K & 5K & 670K & 8K\\ 
% Quora-U &  117K & 3K& 20K & 400K & 8K\\ 
% MSCOCO & 110K & 10K & 40K & 500K & 10K\\ 
% \bottomrule
% \end{tabular}
% }
% \label{table:appendix_dataset}
% \end{table} 

\subsection{Implementation Details}

\smallskip
\noindent\textbf{Implementation of LTSL.} The proposed model LTSL is trained across 8, 32GB NVIDIA V100 GPUs via distributed training and its inference can be run on one GPU. The batch size $B$ is set to be 512 for all the datasets. The parameter $K$ for pseudo paraphrase expansion is set to $5$, $T$ for paraphrase generator update is set to $10$ for all the datasets. We use the BERT-base~\cite{devlin2018bert} to build our gold data selector, with 12-layer transformer blocks, 768-dimension hidden state, 12 attention heads and total 110M parameters. Specifically, we use the pre-trained BERT-Base-Uncased version. For the paraphrase generator, we use BART-large~\cite{lewis2019bart}, a pre-trained seq2seq transformer with 400M parameter. For training stage, we use Adam optimizer for fine-tuning with
$\beta_1$ as 0.9, $\beta_2$ as 0.999. The max sequence length of BERT input is set to 66. For learning LTSL, we train the model for 10000 epochs with early-stopping strategy. We grid search for the learning rate in \{0.0001, 0.0005, 0.001, 0.005, 0.01, 0.05, 0.1\}, L2 regularization in \{$10^{-6}$, $10^{-5}$, $10^{-4}$, $10^{-3}$, $10^{-2}$, $10^{-1}$\} and the dropout rate in \{0.1, 0.2, 0.3, 0.4, 0.5, 0.6, 0.7\}. The optimal values are selected when the model achieves the highest accuracy for the validation samples.

\smallskip
\noindent\textbf{Model Pre-training.}
Pre-training is widely used by reinforcement learning based methods to accelerate the training of RL agent~\cite{yoon2020data,zhao2020reinforced}. Specifically, we use pair-wise ranking loss~\cite{chen2009ranking} to pre-train the gold data selector using the retrieved data from BM25. This way we can make sure the labeled paraphrases will not leak duing the model pre-training stage.
% our base matching model, with 12-layer transformer blocks, 768-dimension hidden state, 12 attention heads and total 110M parameters. We
% use the pre-trained BERT-Base-Uncased∗
% for the
% English benchmarks and BERT-Base-Chinese†
% for
% e-commerce dataset. For training stage, we use
% Adam (Kingma and Ba, 2014) for fine-tuning with
% β1 as 0.9, β2 as 0.999. The max sequence length
% of BERT input is set to 64. For other hyperparameters, we set learning rate as 5e-5, ratio
% δ = size(Ω)/M as 0.2, iteration number N1 as 5
% and episode number N2 as 20. We select weight λ

% \smallskip
% \noindent\textbf{Packages Used for Implementation.} The relevant packages that we use in the implementation and their corresponding versions are as following: python==3.6.6, torch==1.4.0, cuda==10.2, tensorboard==1.10.0, numpy==1.14.5, scipy==1.1.0, NLTK==3.4.5 and scikit-learn==0.21.3.

\smallskip
\noindent\textbf{Implementation of Baselines.}
Since many recent works on paraphrase generation have not released their implementations, we follow the same data pre-process procedure and the train/test split as used in their papers to make a fair comparison. This way for the supervised and unsupervised baseline methods, we can directly adopt the reported in their papers. Specifically, for the comparison on two datasets Quora-S and Twitter, the results of Residual LSTM, Transformer, RbM, RaE and FSTE are from \cite{li2018paraphrase,kazemnejad2020paraphrase}; For the comparison on Quora-U and MSCOCO, we adopt the results of CGMH, UPSA, PUP and set2seq+RTT from \cite{liu2019unsupervised, siddique2020unsupervised,guoautomatically}. For BackTrans, we use the Opus-MT~\cite{tiedemann2020opus} to conduct EN-FR and FR-EN back-translation. For the results of human evaluation (Table \ref{table:human}) and case study (Table \ref{table:case}), we run the implementation of UPSA\footnote{\url{https://github.com/Liuxg16/UPSA}} published with the paper and adopt the same parameters as described in the paper~\cite{liu2019unsupervised}.

%% file: emnlp2021.bbl
\begin{thebibliography}{61}
\expandafter\ifx\csname natexlab\endcsname\relax\def\natexlab#1{#1}\fi

\bibitem[{Aprosio et~al.(2019)Aprosio, Tonelli, Turchi, Negri, and
  Di~Gangi}]{aprosio2019neural}
Alessio~Palmero Aprosio, Sara Tonelli, Marco Turchi, Matteo Negri, and Mattia~A
  Di~Gangi. 2019.
\newblock Neural text simplification in low-resource conditions using weak
  supervision.
\newblock In \emph{ACL NeuralGen Workshop}.

\bibitem[{Bannard and Callison-Burch(2005)}]{bannard2005paraphrasing}
Colin Bannard and Chris Callison-Burch. 2005.
\newblock Paraphrasing with bilingual parallel corpora.
\newblock In \emph{ACL}.

\bibitem[{Bao et~al.(2019)Bao, Zhou, Huang, Li, Mou, Vechtomova, Dai, and
  Chen}]{bao2019generating}
Yu~Bao, Hao Zhou, Shujian Huang, Lei Li, Lili Mou, Olga Vechtomova, Xinyu Dai,
  and Jiajun Chen. 2019.
\newblock Generating sentences from disentangled syntactic and semantic spaces.
\newblock In \emph{ACL}.

\bibitem[{Berant and Liang(2014)}]{berant2014semantic}
Jonathan Berant and Percy Liang. 2014.
\newblock Semantic parsing via paraphrasing.
\newblock In \emph{ACL}.

\bibitem[{Bowman et~al.(2016)Bowman, Vilnis, Vinyals, Dai, Jozefowicz, and
  Bengio}]{bowman2016generating}
Samuel Bowman, Luke Vilnis, Oriol Vinyals, Andrew Dai, Rafal Jozefowicz, and
  Samy Bengio. 2016.
\newblock Generating sentences from a continuous space.
\newblock In \emph{SIGNLL}.

\bibitem[{Chen et~al.(2019{\natexlab{a}})Chen, Tang, Wiseman, and
  Gimpel}]{chen2019controllable}
Mingda Chen, Qingming Tang, Sam Wiseman, and Kevin Gimpel. 2019{\natexlab{a}}.
\newblock Controllable paraphrase generation with a syntactic exemplar.
\newblock In \emph{ACL}.

\bibitem[{Chen et~al.(2019{\natexlab{b}})Chen, Liao, Chen, and
  Zhang}]{chen2019understanding}
Pengfei Chen, Ben~Ben Liao, Guangyong Chen, and Shengyu Zhang.
  2019{\natexlab{b}}.
\newblock Understanding and utilizing deep neural networks trained with noisy
  labels.
\newblock In \emph{ICML}.

\bibitem[{Chen et~al.(2009)Chen, Liu, Lan, Ma, and Li}]{chen2009ranking}
Wei Chen, Tie-Yan Liu, Yanyan Lan, Zhi-Ming Ma, and Hang Li. 2009.
\newblock Ranking measures and loss functions in learning to rank.
\newblock In \emph{NeurIPS}.

\bibitem[{Dehghani et~al.(2017)Dehghani, Zamani, Severyn, Kamps, and
  Croft}]{dehghani2017neural}
Mostafa Dehghani, Hamed Zamani, Aliaksei Severyn, Jaap Kamps, and W~Bruce
  Croft. 2017.
\newblock Neural ranking models with weak supervision.
\newblock In \emph{SIGIR}.

\bibitem[{Devlin et~al.(2018)Devlin, Chang, Lee, and
  Toutanova}]{devlin2018bert}
Jacob Devlin, Ming-Wei Chang, Kenton Lee, and Kristina Toutanova. 2018.
\newblock Bert: Pre-training of deep bidirectional transformers for language
  understanding.
\newblock \emph{arXiv preprint arXiv:1810.04805}.

\bibitem[{Dolan et~al.(2004)Dolan, Quirk, and Brockett}]{dolan2004unsupervised}
Bill Dolan, Chris Quirk, and Chris Brockett. 2004.
\newblock Unsupervised construction of large paraphrase corpora: Exploiting
  massively parallel news sources.
\newblock In \emph{COLING}.

\bibitem[{Dong et~al.(2017)Dong, Mallinson, Reddy, and
  Lapata}]{dong2017learning}
Li~Dong, Jonathan Mallinson, Siva Reddy, and Mirella Lapata. 2017.
\newblock Learning to paraphrase for question answering.
\newblock In \emph{EMNLP}.

\bibitem[{Feng et~al.(2018)Feng, Huang, Zhao, Yang, and
  Zhu}]{feng2018reinforcement}
Jun Feng, Minlie Huang, Li~Zhao, Yang Yang, and Xiaoyan Zhu. 2018.
\newblock Reinforcement learning for relation classification from noisy data.
\newblock In \emph{AAAI}.

\bibitem[{Fu et~al.(2019)Fu, Feng, and Cunningham}]{fu2019paraphrase}
Yao Fu, Yansong Feng, and John~P Cunningham. 2019.
\newblock Paraphrase generation with latent bag of words.
\newblock In \emph{NeurIPS}.

\bibitem[{Ghosh et~al.(2017)Ghosh, Kumar, and Sastry}]{ghosh2017robust}
Aritra Ghosh, Himanshu Kumar, and PS~Sastry. 2017.
\newblock Robust loss functions under label noise for deep neural networks.
\newblock In \emph{AAAI}.

\bibitem[{Goldberger and Ben-Reuven(2017)}]{goldberger2016training}
Jacob Goldberger and Ehud Ben-Reuven. 2017.
\newblock Training deep neural-networks using a noise adaptation layer.
\newblock In \emph{ICLR}.

\bibitem[{Gormley and Tong(2015)}]{gormley2015elasticsearch}
Clinton Gormley and Zachary Tong. 2015.
\newblock Elasticsearch: The definitive guide.

\bibitem[{Guo et~al.(2021)Guo, Huang, Zhu, Chen, Zhang, Chen, and
  Huang}]{guoautomatically}
Zilu Guo, Zhongqiang Huang, Kenny~Q Zhu, Guandan Chen, Kaibo Zhang, Boxing
  Chen, and Fei Huang. 2021.
\newblock Automatically paraphrasing via sentence reconstruction and round-trip
  translation.
\newblock In \emph{IJCAI}.

\bibitem[{Guu et~al.(2018)Guu, Hashimoto, Oren, and Liang}]{guu2018generating}
Kelvin Guu, Tatsunori~B Hashimoto, Yonatan Oren, and Percy Liang. 2018.
\newblock Generating sentences by editing prototypes.
\newblock In \emph{TACL}.

\bibitem[{Hashimoto et~al.(2018)Hashimoto, Guu, Oren, and
  Liang}]{hashimoto2018retrieve}
Tatsunori~B Hashimoto, Kelvin Guu, Yonatan Oren, and Percy~S Liang. 2018.
\newblock A retrieve-and-edit framework for predicting structured outputs.
\newblock In \emph{NeurIPS}.

\bibitem[{Hendrycks et~al.(2018)Hendrycks, Mazeika, Wilson, and
  Gimpel}]{hendrycks2018using}
Dan Hendrycks, Mantas Mazeika, Duncan Wilson, and Kevin Gimpel. 2018.
\newblock Using trusted data to train deep networks on labels corrupted by
  severe noise.
\newblock In \emph{NeurIPS}.

\bibitem[{Huang and Chang(2021)}]{huang2021generating}
Kuan-Hao Huang and Kai-Wei Chang. 2021.
\newblock Generating syntactically controlled paraphrases without using
  annotated parallel pairs.
\newblock In \emph{EACL}.

\bibitem[{Iyyer et~al.(2018)Iyyer, Wieting, Gimpel, and
  Zettlemoyer}]{iyyer2018adversarial}
Mohit Iyyer, John Wieting, Kevin Gimpel, and Luke Zettlemoyer. 2018.
\newblock Adversarial example generation with syntactically controlled
  paraphrase networks.
\newblock In \emph{NAACL}.

\bibitem[{Kazemnejad et~al.(2020)Kazemnejad, Salehi, and
  Baghshah}]{kazemnejad2020paraphrase}
Amirhossein Kazemnejad, Mohammadreza Salehi, and Mahdieh~Soleymani Baghshah.
  2020.
\newblock Paraphrase generation by learning how to edit from samples.
\newblock In \emph{ACL}.

\bibitem[{Lan et~al.(2017)Lan, Qiu, He, and Xu}]{lan2017continuously}
Wuwei Lan, Siyu Qiu, Hua He, and Wei Xu. 2017.
\newblock A continuously growing dataset of sentential paraphrases.
\newblock In \emph{EMNLP}.

\bibitem[{Lewis et~al.(2019)Lewis, Liu, Goyal, Ghazvininejad, Mohamed, Levy,
  Stoyanov, and Zettlemoyer}]{lewis2019bart}
Mike Lewis, Yinhan Liu, Naman Goyal, Marjan Ghazvininejad, Abdelrahman Mohamed,
  Omer Levy, Ves Stoyanov, and Luke Zettlemoyer. 2019.
\newblock Bart: Denoising sequence-to-sequence pre-training for natural
  language generation, translation, and comprehension.
\newblock \emph{arXiv preprint arXiv:1910.13461}.

\bibitem[{Lewis et~al.(2020)Lewis, Perez, Piktus, Petroni, Karpukhin, Goyal,
  K{\"u}ttler, Lewis, Yih, Rockt{\"a}schel et~al.}]{lewis2020retrieval}
Patrick Lewis, Ethan Perez, Aleksandara Piktus, Fabio Petroni, Vladimir
  Karpukhin, Naman Goyal, Heinrich K{\"u}ttler, Mike Lewis, Wen-tau Yih, Tim
  Rockt{\"a}schel, et~al. 2020.
\newblock Retrieval-augmented generation for knowledge-intensive nlp tasks.
\newblock \emph{arXiv preprint arXiv:2005.11401}.

\bibitem[{Li et~al.(2019{\natexlab{a}})Li, Wong, Zhao, and
  Kankanhalli}]{li2019learning}
Junnan Li, Yongkang Wong, Qi~Zhao, and Mohan~S Kankanhalli. 2019{\natexlab{a}}.
\newblock Learning to learn from noisy labeled data.
\newblock In \emph{CVPR}.

\bibitem[{Li et~al.(2017)Li, Yang, Song, Cao, Luo, and Li}]{li2017learning}
Yuncheng Li, Jianchao Yang, Yale Song, Liangliang Cao, Jiebo Luo, and Li-Jia
  Li. 2017.
\newblock Learning from noisy labels with distillation.
\newblock In \emph{ICCV}.

\bibitem[{Li et~al.(2018)Li, Jiang, Shang, and Li}]{li2018paraphrase}
Zichao Li, Xin Jiang, Lifeng Shang, and Hang Li. 2018.
\newblock Paraphrase generation with deep reinforcement learning.
\newblock In \emph{EMNLP}.

\bibitem[{Li et~al.(2019{\natexlab{b}})Li, Jiang, Shang, and
  Liu}]{li2019decomposable}
Zichao Li, Xin Jiang, Lifeng Shang, and Qun Liu. 2019{\natexlab{b}}.
\newblock Decomposable neural paraphrase generation.
\newblock In \emph{ACL}.

\bibitem[{Lin(2004)}]{lin2004rouge}
Chin-Yew Lin. 2004.
\newblock Rouge: A package for automatic evaluation of summaries.
\newblock In \emph{ACL Workshop}.

\bibitem[{Lin et~al.(2014)Lin, Maire, Belongie, Hays, Perona, Ramanan,
  Doll{\'a}r, and Zitnick}]{lin2014microsoft}
Tsung-Yi Lin, Michael Maire, Serge Belongie, James Hays, Pietro Perona, Deva
  Ramanan, Piotr Doll{\'a}r, and C~Lawrence Zitnick. 2014.
\newblock Microsoft coco: Common objects in context.
\newblock In \emph{ECCV}.

\bibitem[{Liu et~al.(2019)Liu, Mou, Meng, Zhou, Zhou, and
  Song}]{liu2019unsupervised}
Xianggen Liu, Lili Mou, Fandong Meng, Hao Zhou, Jie Zhou, and Sen Song. 2019.
\newblock Unsupervised paraphrasing by simulated annealing.
\newblock \emph{arXiv preprint arXiv:1909.03588}.

\bibitem[{McKeown(1983)}]{mckeown1983paraphrasing}
Kathleen McKeown. 1983.
\newblock Paraphrasing questions using given and new information.
\newblock \emph{American Journal of Computational Linguistics}.

\bibitem[{McKeown(1980)}]{mckeown1980paraphrasing}
Kathleen~R McKeown. 1980.
\newblock Paraphrasing using given and new information in a question-answer
  system.
\newblock \emph{Technical Reports (CIS)}.

\bibitem[{Miao et~al.(2019)Miao, Zhou, Mou, Yan, and Li}]{miao2019cgmh}
Ning Miao, Hao Zhou, Lili Mou, Rui Yan, and Lei Li. 2019.
\newblock Cgmh: Constrained sentence generation by metropolis-hastings
  sampling.
\newblock In \emph{AAAI}.

\bibitem[{Mirzasoleiman et~al.(2020)Mirzasoleiman, Cao, and
  Leskovec}]{mirzasoleiman2020coresets}
Baharan Mirzasoleiman, Kaidi Cao, and Jure Leskovec. 2020.
\newblock Coresets for robust training of neural networks against noisy labels.
\newblock In \emph{NeurIPS}.

\bibitem[{Niu et~al.(2020)Niu, Yavuz, Zhou, Wang, Keskar, and
  Xiong}]{niu2020unsupervised}
Tong Niu, Semih Yavuz, Yingbo Zhou, Huan Wang, Nitish~Shirish Keskar, and
  Caiming Xiong. 2020.
\newblock Unsupervised paraphrase generation via dynamic blocking.
\newblock \emph{arXiv preprint arXiv:2010.12885}.

\bibitem[{Papineni et~al.(2002)Papineni, Roukos, Ward, and
  Zhu}]{papineni2002bleu}
Kishore Papineni, Salim Roukos, Todd Ward, and Wei-Jing Zhu. 2002.
\newblock Bleu: a method for automatic evaluation of machine translation.
\newblock In \emph{ACL}.

\bibitem[{Patrini et~al.(2017)Patrini, Rozza, Krishna~Menon, Nock, and
  Qu}]{patrini2017making}
Giorgio Patrini, Alessandro Rozza, Aditya Krishna~Menon, Richard Nock, and
  Lizhen Qu. 2017.
\newblock Making deep neural networks robust to label noise: A loss correction
  approach.
\newblock In \emph{CVPR}.

\bibitem[{Post(2018)}]{post2018call}
Matt Post. 2018.
\newblock A call for clarity in reporting bleu scores.
\newblock \emph{arXiv preprint arXiv:1804.08771}.

\bibitem[{Prakash et~al.(2016)Prakash, Hasan, Lee, Datla, Qadir, Liu, and
  Farri}]{prakash2016neural}
Aaditya Prakash, Sadid~A Hasan, Kathy Lee, Vivek Datla, Ashequl Qadir, Joey
  Liu, and Oladimeji Farri. 2016.
\newblock Neural paraphrase generation with stacked residual lstm networks.
\newblock In \emph{COLING}.

\bibitem[{Qin et~al.(2018)Qin, Xu, and Wang}]{qin2018robust}
Pengda Qin, Weiran Xu, and William~Yang Wang. 2018.
\newblock Robust distant supervision relation extraction via deep reinforcement
  learning.
\newblock In \emph{ACL}.

\bibitem[{Reed et~al.(2014)Reed, Lee, Anguelov, Szegedy, Erhan, and
  Rabinovich}]{reed2014training}
Scott Reed, Honglak Lee, Dragomir Anguelov, Christian Szegedy, Dumitru Erhan,
  and Andrew Rabinovich. 2014.
\newblock Training deep neural networks on noisy labels with bootstrapping.
\newblock \emph{arXiv preprint arXiv:1412.6596}.

\bibitem[{Ren et~al.(2018)Ren, Zeng, Yang, and Urtasun}]{ren2018learning}
Mengye Ren, Wenyuan Zeng, Bin Yang, and Raquel Urtasun. 2018.
\newblock Learning to reweight examples for robust deep learning.
\newblock In \emph{ICML}.

\bibitem[{Ren et~al.(2020)Ren, Li, Su, Kartchner, Mitchell, and
  Zhang}]{ren2020denoising}
Wendi Ren, Yinghao Li, Hanting Su, David Kartchner, Cassie Mitchell, and Chao
  Zhang. 2020.
\newblock Denoising multi-source weak supervision for neural text
  classification.
\newblock In \emph{EMNLP}.

\bibitem[{Resnik et~al.(2010)Resnik, Buzek, Hu, Kronrod, Quinn, and
  Bederson}]{resnik2010improving}
Philip Resnik, Olivia Buzek, Chang Hu, Yakov Kronrod, Alex Quinn, and
  Benjamin~B Bederson. 2010.
\newblock Improving translation via targeted paraphrasing.
\newblock In \emph{EMNLP}.

\bibitem[{Robertson and Zaragoza(2009)}]{robertson2009probabilistic}
Stephen Robertson and Hugo Zaragoza. 2009.
\newblock \emph{The probabilistic relevance framework: BM25 and beyond}.
\newblock Now Publishers Inc.

\bibitem[{Siddique et~al.(2020)Siddique, Oymak, and
  Hristidis}]{siddique2020unsupervised}
AB~Siddique, Samet Oymak, and Vagelis Hristidis. 2020.
\newblock Unsupervised paraphrasing via deep reinforcement learning.
\newblock In \emph{KDD}.

\bibitem[{Sun and Zhou(2012)}]{sun2012joint}
Hong Sun and Ming Zhou. 2012.
\newblock Joint learning of a dual smt system for paraphrase generation.
\newblock In \emph{ACL}.

\bibitem[{Tanaka et~al.(2018)Tanaka, Ikami, Yamasaki, and
  Aizawa}]{tanaka2018joint}
Daiki Tanaka, Daiki Ikami, Toshihiko Yamasaki, and Kiyoharu Aizawa. 2018.
\newblock Joint optimization framework for learning with noisy labels.
\newblock In \emph{CVPR}.

\bibitem[{Tiedemann and Thottingal(2020)}]{tiedemann2020opus}
J{\"o}rg Tiedemann and Santhosh Thottingal. 2020.
\newblock Opus-mt--building open translation services for the world.
\newblock In \emph{EAMT}.

\bibitem[{Vaswani et~al.(2017)Vaswani, Shazeer, Parmar, Uszkoreit, Jones,
  Gomez, Kaiser, and Polosukhin}]{vaswani2017attention}
Ashish Vaswani, Noam Shazeer, Niki Parmar, Jakob Uszkoreit, Llion Jones,
  Aidan~N Gomez, {\L}ukasz Kaiser, and Illia Polosukhin. 2017.
\newblock Attention is all you need.
\newblock In \emph{NeurIPS}.

\bibitem[{Wieting et~al.(2017)Wieting, Mallinson, and
  Gimpel}]{wieting2017learning}
John Wieting, Jonathan Mallinson, and Kevin Gimpel. 2017.
\newblock Learning paraphrastic sentence embeddings from back-translated
  bitext.
\newblock In \emph{EMNLP}.

\bibitem[{Williams(1992)}]{williams1992simple}
Ronald~J Williams. 1992.
\newblock Simple statistical gradient-following algorithms for connectionist
  reinforcement learning.
\newblock \emph{Machine learning}.

\bibitem[{Yao et~al.(2019)Yao, Peddamail, and Sun}]{yao2019coacor}
Ziyu Yao, Jayavardhan~Reddy Peddamail, and Huan Sun. 2019.
\newblock Coacor: Code annotation for code retrieval with reinforcement
  learning.
\newblock In \emph{The Web Conference}.

\bibitem[{Yoon et~al.(2020)Yoon, Arik, and Pfister}]{yoon2020data}
Jinsung Yoon, Sercan Arik, and Tomas Pfister. 2020.
\newblock Data valuation using reinforcement learning.
\newblock In \emph{ICML}.

\bibitem[{Zhang et~al.(2020)Zhang, Zhang, Arik, Lee, and
  Pfister}]{zhang2020distilling}
Zizhao Zhang, Han Zhang, Sercan~O Arik, Honglak Lee, and Tomas Pfister. 2020.
\newblock Distilling effective supervision from severe label noise.
\newblock In \emph{CVPR}.

\bibitem[{Zhao et~al.(2020)Zhao, Wu, Niu, and Wang}]{zhao2020reinforced}
Mingjun Zhao, Haijiang Wu, Di~Niu, and Xiaoli Wang. 2020.
\newblock Reinforced curriculum learning on pre-trained neural machine
  translation models.
\newblock \emph{arXiv preprint arXiv:2004.05757}.

\bibitem[{Zheng et~al.(2021)Zheng, {Hassan Awadallah}, and
  Dumais}]{zheng2021meta}
Guoqing Zheng, Ahmed {Hassan Awadallah}, and Susan Dumais. 2021.
\newblock Meta label correction for noisy label learning.
\newblock In \emph{AAAI}.

\end{thebibliography}
